\documentclass[11pt, a4paper]{article}

\usepackage[utf8]{inputenc}
\usepackage[T1]{fontenc}
\usepackage{geometry}
\usepackage{amsmath, amssymb, amsfonts}
\usepackage{graphicx}
\usepackage{booktabs} % For professional tables
\usepackage{hyperref} % For hyperlinks
\usepackage{xcolor}   % For coloring text
\usepackage{setspace} % For line spacing
\usepackage{titlesec} % For custom section formatting
\usepackage{algorithmic} % For custom section formatting
\usepackage{algorithm} % For custom section formatting
\usepackage{float}
\usepackage{bm}
\usepackage[nopatch=footnote]{microtype}

\hypersetup{
    colorlinks=true,
    linkcolor=blue,
    citecolor=blue,
    urlcolor=blue
}

\title{\textbf{Robust X-Learner: Breaking the Curse of Imbalance and Heavy Tails via Robust Cross-Imputation}}
\author{
    \textbf{Eichi Uehara} \\
    CEO, Aflo Technologies, Inc. \\
    \texttt{eichi.uehara@aflo.one}
}
\date{January 2026}

\begin{document}

\maketitle

\begin{abstract}
\noindent Estimating Heterogeneous Treatment Effects (HTE) in industrial applications such as 
AdTech and healthcare presents a dual challenge: extreme class imbalance and heavy-tailed outcome distributions. While the X-Learner framework (Künzel et al., 2019) effectively addresses imbalance through cross-imputation, we demonstrate that it is fundamentally vulnerable to \textit{Outlier Smearing} when reliant on Mean Squared Error (MSE) minimization. In this failure mode, the bias from a few extreme observations (``whales'') in the minority group is propagated to the entire majority group during the imputation step, corrupting the estimated treatment effect structure. To resolve this, we propose the \textbf{Robust X-Learner (RX-Learner)}. This framework integrates a redescending $\gamma$-divergence objective—structurally equivalent to the Welsch loss under Gaussian assumptions—into the gradient boosting machinery. We further stabilize the non-convex optimization using a Proxy Hessian strategy grounded in Majorization-Minimization (MM) principles. Empirical evaluation on a semi-synthetic Criteo Uplift dataset \cite{criteo2018} demonstrates that the RX-Learner reduces the Precision in Estimation of Heterogeneous Effect (PEHE) \textbf{metric} by \textbf{98.6\%} compared to the standard X-Learner, effectively decoupling the stable ``Core'' population from the volatile ``Periphery.''
\end{abstract}

\newpage

\section{Introduction}

\subsection{Background: The Shift from ATE to CATE}
In the era of large-scale observational data, the focus of causal inference has decisively shifted from estimating the Average Treatment Effect (ATE) to understanding the Conditional Average Treatment Effect (CATE). In domains ranging from precision medicine to digital marketing, decision-makers require granular insights: identifying which patient subgroups respond to a specific drug, or which user segments yield incremental lift from an advertisement \cite{ate_to_cate}.

However, the transition from academic theory to industrial practice is often obstructed by the inherent messiness of real-world data generating processes (DGP). While pioneering approaches like recursive partitioning \cite{ate_to_cate} have successfully addressed the challenge of high-dimensional covariates, industrial data often exhibits specific pathologies—such as heavy-tailed noise and extreme outliers—that violate the sub-Gaussian assumptions typically required for these estimators to function optimally.

\subsection{The Dual Challenge: Imbalance and Heavy Tails}
We identify two distinct but often co-occurring barriers to effective CATE estimation in modern applications:

\begin{enumerate}
    \item \textbf{Extreme Imbalance (The $N_1 \ll N_0$ Problem):} In scenarios such as retargeting advertising or rare-disease clinical trials, the size of the treatment group ($N_1$) is often orders of magnitude smaller than the control group ($N_0$), or vice-versa. Standard estimators like the T-Learner \cite{xlearner} fail here because the variance of the CATE estimator is dominated by the sample size of the smaller group, effectively wasting the information available in the larger group.
    
    \item \textbf{Heavy Tails and ``Whales'':} Outcomes such as customer lifetime value (LTV), mobile game monetization, or medical costs typically follow a Pareto-like distribution. A minute fraction of observations—colloquially termed ``whales''—account for the vast majority of the cumulative value. Standard machine learning models trained via Mean Squared Error (MSE) minimization are hypersensitive to these values. The squared error penalty forces the model to prioritize fitting the outliers at the expense of the structural majority (the ``Core'' population).
\end{enumerate}

While the X-Learner was explicitly designed to handle the first challenge (Imbalance) by employing a cross-imputation strategy \cite{xlearner}, it remains defenseless against the second (Heavy Tails).

\subsection{The Failure of Existing Methods: Outlier Smearing}
A naive approach might suggest using Robust Double Machine Learning (DML) \cite{dml}, R-Learners \cite{rlearner}, or tree-based methods like Causal Forests \cite{ate_to_cate}. However, standard splitting criteria in causal trees are typically based on MSE reductions, which remain sensitive to extreme values. Furthermore, R-Learners rely on inverse probability weighting (IPW), which becomes numerically unstable when the overlap assumption is threatened by extreme imbalance (i.e., propensity scores $\pi(x)$ approach 0 or 1).

Conversely, the standard X-Learner handles imbalance well but fails catastrophically under heavy tails due to a phenomenon we term \textbf{Outlier Smearing}.

Consider the first stage of the X-Learner, where response functions $\mu_1(x)$ and $\mu_0(x)$ are estimated. If the treatment group is small and contains even a single outlier, an MSE-based $\hat{\mu}_1(x)$ will be significantly biased upward. In the subsequent imputation step, the pseudo-outcome for the control group is calculated as:

\begin{equation}
    \tilde{D}^0_i = \hat{\mu}_1(X_i) - Y_i \quad \forall i \in \text{Control Group}.
    \label{eq:imputation}
\end{equation}

Because $\hat{\mu}_1(x)$ is contaminated globally or locally by a few whales, this bias is effectively ``smeared'' across every imputed value in the control group. Consequently, the final CATE estimator learns a distorted structure, mistaking the noise of the outlier for a genuine treatment effect, even in regions where the control data is dense and clean.

\subsection{The Proposed Method: Robust X-Learner}
To overcome this dilemma, we propose the \textbf{Robust X-Learner (RX-Learner)}, a unified framework that preserves the structural advantages of the X-Learner while enforcing robustness against heavy tails.

Our approach rests on the concept of \textit{Core-Periphery Causal Inference}. We posit that the data consists of a stable ``Core'' mechanism and a volatile ``Periphery'' (tail). Our goal is to estimate the causal effect on the Core without being misled by the Periphery. The RX-Learner achieves this through three key technical innovations:

\begin{enumerate}
    \item \textbf{Robust Base Learners via $\gamma$-Divergence:} We replace MSE with a density-power divergence objective \cite{fujisawa}. We show that under standard Gaussian assumptions, this objective is mathematically equivalent to the Welsch (or Leclerc) loss used in robust statistics \cite{holland1977}. This loss function possesses a \textit{redescending influence function}, meaning that the weight assigned to extreme outliers naturally decays to zero, effectively performing an ``Oracle Refinement'' of the data during training.
    
    \item \textbf{MM-Based Optimization (Proxy Hessian):} Since the resulting robust objective is non-convex, standard gradient boosting implementations may face stability issues. We introduce a \textbf{Proxy Hessian} technique. Rather than a heuristic approximation, we ground this approach in the Majorization-Minimization (MM) framework \cite{hunter2004}, ensuring that the optimization steps guarantee a monotonic decrease in the loss function even in the presence of extreme contamination.
    
    \item \textbf{Robust Cross-Imputation \& Aggregation:} By ensuring that the base models $\hat{\mu}_1$ and $\hat{\mu}_0$ are robust to outliers, we prevent the Smearing effect at its source. The subsequent aggregation step uses a corrected inverse-variance weighting scheme to further insulate the estimate from high-variance regions.
\end{enumerate}

We validate the RX-Learner on a semi-synthetic derivation of the Criteo Uplift v2.1 dataset. Results demonstrate that our method reduces the error on the Core population by \textbf{98.6\%} compared to the baseline, confirming that robustification is essential for industrial-scale CATE estimation.

The remainder of this paper is organized as follows: Section 2 details the mathematical preliminaries. Section 3 analyzes the failure modes of DML and standard X-Learners. Section 4 presents the RX-Learner algorithm. Section 5 provides simulation results demonstrating the method's superiority, and Section 6 discusses the application to AdTech data.

\section{Preliminaries and Problem Setup}

In this Section, we establish the theoretical framework for estimating Conditional Average Treatment Effects (CATE). We first review the standard potential outcomes framework and the specific mechanics of the original X-Learner. Subsequently, we formally define the ``Dual Challenge'' of data asymmetry and heavy-tailed contamination, utilizing a mixture model to characterize the ``Core-Periphery'' structure of industrial data. Finally, we discuss the theoretical properties of robust loss functions relevant to our proposed solution.

\subsection{Potential Outcomes and CATE}
We adopt the Neyman-Rubin potential outcomes framework \cite{rubin1974}. We consider a set of $N$ independent and identically distributed (i.i.d.) units. For each unit $i$, we observe a feature vector $X_i \in \mathcal{X} \subseteq \mathbb{R}^d$, a binary treatment assignment $W_i \in \{0, 1\}$, and an observed outcome $Y_i \in \mathbb{R}$.

We posit the existence of two potential outcomes: $Y_i(1)$ corresponding to the treatment condition, and $Y_i(0)$ corresponding to the control condition. The observed outcome is realized as:
\begin{equation}
    Y_i = W_i Y_i(1) + (1 - W_i) Y_i(0)
\end{equation}

Our primary estimand is the Conditional Average Treatment Effect (CATE), defined as:
\begin{equation}
    \tau(x) = \mathbb{E}[Y(1) - Y(0) \mid X = x] = \mu_1(x) - \mu_0(x)
\end{equation}
where $\mu_w(x) = \mathbb{E}[Y(w) \mid X = x]$ represents the response function for the treatment arm $w \in \{0, 1\}$.

To identify $\tau(x)$ from observational data, we invoke the standard assumptions of Causal Inference:
\begin{enumerate}
    \item \textbf{Unconfoundedness:} $\{Y(1), Y(0)\} \perp W \mid X$. This implies that treatment assignment is random conditional on the observed covariates.
    \item \textbf{Overlap (Positivity):} There exists $\eta > 0$ such that $\eta < \pi(x) < 1 - \eta$ for all $x \in \mathcal{X}$, where $\pi(x) = P(W = 1 \mid X = x)$ is the propensity score.
\end{enumerate}

\subsection{The Standard X-Learner Algorithm}
The X-Learner, introduced by Künzel et al. (2019) \cite{xlearner}, is a meta-learner specifically designed to maximize information utilization when sample sizes are unbalanced. It proceeds in three steps:

\begin{itemize}
    \item \textbf{Step 1 (Response Estimation):} Estimate the response functions $\mu_1(x)$ and $\mu_0(x)$ using base learners (e.g., Random Forest, XGBoost) trained on the treated and control groups, respectively. Let these estimates be $\hat{\mu}_1$ and $\hat{\mu}_0$.
    
    \item \textbf{Step 2 (Imputation):} Impute the unobserved counterfactuals to generate \textit{pseudo-outcomes} (individualized treatment effects) $\tilde{D}$:
    \begin{align}
        \tilde{D}^1_i &= Y_i - \hat{\mu}_0(X_i) \quad \forall i \in \{i : W_i = 1\} \\
        \tilde{D}^0_i &= \hat{\mu}_1(X_i) - Y_i \quad \forall i \in \{i : W_i = 0\}
    \end{align}
    Note that $\tilde{D}^1$ uses the observed treated outcome and the estimated control baseline, while $\tilde{D}^0$ uses the estimated treated response and the observed control outcome.
    
    \item \textbf{Step 3 (Aggregation):} Train two models $\hat{\tau}_1(x)$ and $\hat{\tau}_0(x)$ to predict $\tilde{D}^1$ and $\tilde{D}^0$ respectively. The final CATE estimate is a weighted average:
    \begin{equation}
        \hat{\tau}(x) = g(x)\hat{\tau}_0(x) + (1 - g(x))\hat{\tau}_1(x)
    \end{equation}
    where $g(x) \in [0, 1]$ is a weighting function, typically chosen as the estimated propensity score $\hat{\pi}(x)$ to assign more weight to the estimator trained on the larger group (conceptually).
\end{itemize}

Standard implementations assume that the base learners in Steps 1 and 3 minimize the Mean Squared Error (MSE), implicitly assuming Gaussian residuals with constant variance.

\subsection{Problem Formulation: The Dual Challenge}
Real-world data, particularly in AdTech and healthcare, violates the implicit assumptions of the standard X-Learner. We formally define the two specific pathologies addressed in this paper.

\subsubsection{Pathology 1: Extreme Imbalance}
We consider the regime where the sample sizes of the treatment and control groups are highly asymmetric. Without loss of generality, assume the treatment group is the minority:
\begin{equation}
    \rho = \frac{N_1}{N} \ll 0.5 \quad (\text{e.g., } \rho \approx 0.01 \sim 0.05)
\end{equation}
In this setting, the response function $\hat{\mu}_1(x)$ is estimated with high variance due to data sparsity, while $\hat{\mu}_0(x)$ is estimated with high precision.

\subsubsection{Pathology 2: Heavy-Tailed Contamination (Whales)}
We relax the assumption of sub-Gaussian noise. Instead, we introduce a \textbf{Core-Periphery Contamination Model}. We model the observed outcome $Y$ as being generated from a mixture distribution:
\begin{equation}
    Y_i(w) = \mu_w(X_i) + \varepsilon_i
\end{equation}
where the error term $\varepsilon_i$ is drawn from a contaminated density:
\begin{equation}
    f_\varepsilon(e) = (1 - \alpha)\phi_{\text{core}}(e) + \alpha h_{\text{tail}}(e)
\end{equation}
\begin{itemize}
    \item $\phi_{\text{core}}$ represents the distribution of the ``Core'' population (e.g., $\mathcal{N}(0, \sigma^2)$).
    \item $h_{\text{tail}}$ represents the ``Periphery'' or ``Whale'' distribution (e.g., a Cauchy or Pareto distribution with tail index $\xi > 1$), capable of producing extreme outliers $|y_{\text{out}}| \gg 3\sigma$.
    \item $\alpha \in [0, 1)$ is the contamination rate.
\end{itemize}

Under this model, the standard MSE objective function becomes dominated by the tail component $h_{\text{tail}}$. A single observation from the tail can induce an arbitrarily large gradient update, shifting the estimated mean $\hat{\mu}$ away from the Core expectation.

\subsection{Theoretical Foundation: Robust Losses and Welsch Equivalence}
To address Pathology 2, we require a loss function that is robust to the Periphery component. In this work, we focus on the family of density-power divergences.

Specifically, the $\gamma$-divergence \cite{fujisawa} is known to provide robust parameter estimation by minimizing the discrepancy between a power of the model density and the data density. While generally applicable, its implementation in gradient boosting requires specification of the base distribution.

\textbf{Connection to Welsch Loss:}
If we assume the underlying Core distribution is Gaussian, $\phi_{\text{core}} \sim \mathcal{N}(\mu, \sigma^2)$, the minimization of the $\gamma$-divergence mathematically collapses to minimizing the following loss function with respect to the residual $r_i = y_i - \hat{y}_i$:
\begin{equation}
    \mathcal{L}_{\text{Robust}}(r_i) \propto 1 - \exp\left( - \frac{\gamma r_i^2}{2\sigma^2} \right)
\end{equation}
This formulation is functionally equivalent to the \textbf{Welsch (or Leclerc) Loss} used in robust statistics \cite{holland1977}.

Crucially, this loss function is non-convex and redescending. Its influence function $\psi(r) = \partial \mathcal{L} / \partial r$ satisfies:
\begin{equation}
    \lim_{|r| \to \infty} \psi(r) = 0
\end{equation}
This property effectively performs an ``Oracle Refinement'': residuals that are statistically impossible under the Core distribution (i.e., Whales) are assigned near-zero weight, allowing the model to fit the Core structure without distortion. In Section 4, we will demonstrate how to optimize this non-convex objective stably within a boosting framework using MM algorithms \cite{hunter2004}.

\section{Why Robust DML Fails in Extreme Imbalance}

In this Section, we theoretically analyze why existing state-of-the-art methods are insufficient for the dual challenge of imbalance and heavy tails. We first show that Double Machine Learning (DML) \cite{dml}, even when equipped with robust loss functions, suffers from variance explosion due to propensity score instability. We then dissect the standard X-Learner to demonstrate the mechanism of ``Outlier Smearing,'' proving that blind information borrowing becomes pollution spreading in the presence of whales.

\subsection{The Instability of Robust DML under Asymmetry}
Double Machine Learning (DML), specifically the R-Learner \cite{rlearner}, frames CATE estimation as a weighted regression problem based on the Robinson transformation \cite{robinson1988}. The objective function is typically defined as finding $\tau(x)$ that minimizes:
\begin{equation}
    \hat{\tau}_{\text{DML}} = \arg \min_{\tau} \sum_{i=1}^{N} \left( (Y_i - \hat{\mu}(X_i)) - \tau(X_i)(W_i - \hat{\pi}(X_i)) \right)^2
\end{equation}
where $\hat{\mu}(x) = \mathbb{E}[Y|X=x]$ and $\hat{\pi}(x) = P(W=1|X=x)$.

While DML is efficient under standard conditions, its reliance on the variation of treatment assignment creates a fundamental vulnerability in highly imbalanced settings (e.g., $N_1 \ll N_0$).

\subsubsection{Variance Explosion Mechanism}
The effective variance of the R-Learner estimator scales inversely with the variance of the treatment assignment. Theoretically, the asymptotic variance bound involves a term proportional to:
\begin{equation}
    \text{Var}(\hat{\tau}(x)) \propto \frac{\sigma^2}{\pi(x)(1 - \pi(x))}
\end{equation}
In our setting of extreme imbalance, $\pi(x) \to 0$ for the vast majority of the covariate space $\mathcal{X}$. This leads to a denominator approaching zero, causing the variance of the estimator to explode.

A common counter-argument is to replace the squared loss with a robust loss function $\rho(\cdot)$ (e.g., Huber \cite{huber1964} or Quantile loss) to handle outliers. However, this \textbf{Robust DML} approach fails to address the structural issue. Even with a robust loss, the identification of $\tau(x)$ relies on the variation of $W$ after conditioning on $X$. When $\pi(x) \approx 0$, there is almost no variation in $W$ to exploit. The estimator effectively attempts to divide a robustified residual by a near-zero weight, resulting in numerical instability and wide confidence intervals that render the estimate useless for business decision-making.

\subsection{The Anatomy of Outlier Smearing in Standard X-Learner}
The X-Learner is structurally superior for imbalance because it does not divide by the propensity score. Instead, it utilizes \textit{Cross-Imputation} to borrow strength from the control group to estimate the treated group's effect, and vice versa.

However, we prove here that without robust base learners, this strength-borrowing mechanism facilitates the propagation of errors—a phenomenon we term \textbf{Outlier Smearing}.

\subsubsection{Mathematical Derivation of Smearing}
Consider the standard X-Learner Step 1 (Response Estimation). Let the treatment group ($W=1$) be the minority with sample size $N_1$, containing a single extreme outlier (whale) at index $k$ with value:
\begin{equation}
    y_k = \mu_1(X_k) + \xi, \quad \text{where } \xi \gg 3\sigma \text{ and } W_k=1.
\end{equation}
The base learner $\hat{\mu}_1$, trained via MSE minimization on the small sample $N_1$, will be heavily biased towards this outlier. For a local region $\mathcal{N}(X_k)$ around the outlier, the estimated response is shifted:
\begin{equation}
    \hat{\mu}_1(x) \approx \mu_1(x) + \delta(\xi, N_1) \quad \text{for } x \in \mathcal{N}(X_k)
\end{equation}
where $\delta > 0$ is the bias induced by the whale. Note that because $N_1$ is small, the leverage of the single outlier is high, making $\delta$ significant.

Now, consider Step 2 (Imputation) for the \textbf{Control Group} (the majority, $N_0$). The pseudo-outcome $\tilde{D}^0$ is calculated as:
\begin{equation}
    \tilde{D}^0_i = \hat{\mu}_1(X_i) - Y_i \quad \forall i \in \{i : W_i = 0\} \cap \mathcal{N}(X_k)
\end{equation}
Substituting the biased estimator $\hat{\mu}_1$:
\begin{align}
    \tilde{D}^0_i &\approx (\mu_1(X_i) + \delta) - Y_i \\
    &= (\mu_1(X_i) - Y_i) + \delta
\end{align}
\textbf{The Smearing Effect:} Here lies the critical failure. The bias $\delta$, originating from a single unit in the minority group, is added as a constant shift to \textit{every single imputed data point} in the majority group within that region.

\subsubsection{Global Contamination via Aggregation}
In Step 3, the X-Learner aggregates these pseudo-outcomes. The model $\hat{\tau}_0(x)$ is trained to predict $\tilde{D}^0$. Since the target variable $\tilde{D}^0$ systematically contains the term $+\delta$, the resulting CATE estimate converges to:
\begin{equation}
    \hat{\tau}_0(x) \to \tau(x) + \delta
\end{equation}
Even though the control group $N_0$ is large and clean, the imputed labels provided to it are corrupted. This is ``Smearing'': the pollution from the minority tail is smeared across the clean majority structure.

While the final weighting $g(x)\hat{\tau}_0 + (1 - g(x))\hat{\tau}_1$ usually down-weights $\hat{\tau}_0$ when $\pi(x)$ is small, in practice:
\begin{enumerate}
    \item Propensity scores are rarely exactly zero, allowing the bias $\delta$ to leak into the final estimate.
    \item The alternative estimator $\hat{\tau}_1$ (trained on the treated group) fits the raw outlier $\xi$ directly (since $\tilde{D}^1 = Y_{\text{whale}} - \hat{\mu}_0$), creating a localized spike in the CATE function.
\end{enumerate}

\subsection{Summary of Analytical Comparison}
Table 1 summarizes the theoretical vulnerabilities of each method.

\begin{table}[h]
    \centering
    \caption{Theoretical comparison of CATE estimators under Imbalance and Heavy Tails.}
    \label{tab:comparison}
    \begin{tabular}{lcc}
        \toprule
        \textbf{Method} & \textbf{Weakness under Imbalance} & \textbf{Weakness under Heavy Tails} \\
        \midrule
        T-Learner & High Variance (Unused Data) & Local Overfitting \\
        Robust DML & Variance Explosion ($1/\pi(x)$) & Stable (if overlap exists) \\
        Standard X-Learner & Stable (Information Borrowing) & \textbf{Outlier Smearing} \\
        \textbf{RX-Learner (Ours)} & \textbf{Stable} & \textbf{Robust (Oracle Refinement)} \\
        \bottomrule
    \end{tabular}
\end{table}

This analysis confirms that no existing method solves both problems simultaneously. We require a method that borrows information (like X-Learner) but rejects pollution (like Robust Statistics). This motivates the design of the RX-Learner in the next Section.

\section{The Proposed Method: Robust X-Learner}

In this Section, we present the \textbf{Robust X-Learner (RX-Learner)}, a novel framework designed to estimate CATE consistently in the presence of extreme imbalance and heavy-tailed contamination.

The RX-Learner preserves the structural advantage of the standard X-Learner (Cross-Imputation) but fundamentally alters the learning objective. By replacing Mean Squared Error (MSE) minimization with a robust divergence minimization and employing an MM-based optimization strategy, we achieve an ``End-to-End'' robustness that prevents the Smearing effect identified in Section 3.

\subsection{Step 1: Robust Base Learners via \texorpdfstring{$\gamma$}{Gamma}-Boosting}
The first critical step is to estimate the response functions $\mu_1(x)$ and $\mu_0(x)$ without bias from outliers.

\subsubsection{Objective Function: \texorpdfstring{$\gamma$}{Gamma}-Divergence (Welsch Loss)}
As established in Section 2.4, we adopt the $\gamma$-divergence under a Gaussian assumption. For a regression setting with residuals $r_i = y_i - f(x_i)$, we minimize the empirical loss:
\begin{equation}
    \mathcal{L}_\gamma(\theta) = - \frac{1}{\gamma} \sum_{i=1}^{N} \exp \left( - \frac{\gamma (y_i - f(x_i; \theta))^2}{2\hat{\sigma}^2} \right)
\end{equation}
where $\gamma > 0$ is a tuning parameter controlling robustness (typically $\gamma \in [0.1, 1.0]$), and $\hat{\sigma}$ is a scale parameter.

\subsubsection{The Mechanism of Oracle Refinement}
The gradient of this loss function with respect to the prediction $F = f(x_i)$ reveals the mechanism of robustness:
\begin{equation}
    \frac{\partial \mathcal{L}_\gamma}{\partial F} = - \sum_{i=1}^{N} w_i(r_i) \cdot (y_i - F)
\end{equation}
where the adaptive weight $w_i(r_i)$ is given by:
\begin{equation}
    w_i(r_i) = \exp \left( - \frac{\gamma r_i^2}{2\hat{\sigma}^2} \right)
\end{equation}
\textbf{The Redescending Property:} This weight function $w_i$ is bell-shaped. As the residual magnitude $|r_i| \to \infty$ (i.e., for a ``whale''), the weight $w_i \to 0$ exponentially. This creates an \textit{Oracle Refinement} effect: the model automatically identifies and ``soft-trims'' outliers during the training process, fitting the function solely to the ``Core'' population. Unlike MSE, where outliers pull the fit, here outliers are effectively invisible to the gradient.

\subsubsection{Scale Estimation to Prevent Singularity}
Robust estimators can suffer from ``implosion'' where $\hat{\sigma} \to 0$ to fit a single point perfectly. To prevent this, we employ a robust fixed-scale approach. We initialize $\hat{\sigma}$ using the Median Absolute Deviation (MAD) of the residuals from a robust preliminary fit (e.g., Least Absolute Deviation) \cite{hampel1986}:
\begin{equation}
    \hat{\sigma} = 1.4826 \cdot \text{Median}(|r_i - \text{Median}(r)|)
\end{equation}
This scale serves as a fixed ``anchor,'' ensuring the definition of the ``Core'' remains stable throughout boosting.

\subsection{Optimization: MM-Based Boosting Strategy}
Standard Gradient Boosting Decision Tree (GBDT) implementations (e.g., XGBoost) require convex objectives to utilize Newton-Raphson updates. Since our loss $\mathcal{L}_\gamma$ is non-convex, the Hessian can be negative, leading to instability.

Instead of a heuristic approximation, we ground our optimization in the \textbf{Majorization-Minimization (MM)} framework \cite{idier2001}.

\subsubsection{Quadratic Majorization}
The Welsch loss function $\rho(r) = 1 - e^{-r^2}$ is a \textit{half-quadratic} function \cite{geman1992}. It admits a quadratic majorizer $Q(r; r^{(t)})$ at any current residual $r^{(t)}$:
\begin{equation}
    \mathcal{L}_\gamma(r) \le Q(r; r^{(t)}) = \text{const.} + \frac{1}{2} w_i(r^{(t)}) \cdot r^2
\end{equation}
Minimizing this quadratic upper bound $Q$ corresponds to solving a weighted least squares problem with fixed weights $w_i = w_i(r^{(t)})$.

\subsubsection{The Algorithm}
Based on this principle, we implement the boosting step as follows. For iteration $t$:
\begin{enumerate}
    \item Calculate residuals $r_i^{(t)}$ and weights $w_i^{(t)} = \exp(-\gamma (r_i^{(t)})^2 / 2\hat{\sigma}^2)$.
    \item Fit a regression tree $h_t(x)$ to the gradients, using $w_i^{(t)}$ as the instance weights. This is equivalent to setting the ``Proxy Hessian'' $\tilde{h}_i = w_i^{(t)}$.
    \item Update the model: $F^{(t+1)}(x) \leftarrow F^{(t)}(x) + \eta h_t(x)$.
\end{enumerate}

\textbf{Theorem (Monotone Descent):} Since the update minimizes the majorizing function $Q$, and $Q(\theta, \theta) = \mathcal{L}_\gamma(\theta)$, the procedure guarantees that the objective function decreases monotonically: $\mathcal{L}_\gamma(\theta^{(t+1)}) \le \mathcal{L}_\gamma(\theta^{(t)})$. This ensures stability even without a convex loss surface.

\subsection{Step 2: Robust Cross-Imputation}
With robust base learners $\hat{\mu}_{1,\gamma}$ and $\hat{\mu}_{0,\gamma}$ obtained, we proceed to imputation.
\begin{align}
    \tilde{D}^1_i &= Y_i - \hat{\mu}_{0,\gamma}(X_i) \\
    \tilde{D}^0_i &= \hat{\mu}_{1,\gamma}(X_i) - Y_i
\end{align}
\textbf{Prevention of Smearing:} Since $\hat{\mu}_{1,\gamma}$ effectively ignored the whales in the training phase (due to $w_i \to 0$), the bias term $\delta$ (identified in Section 3) is now 0. Thus, the pseudo-outcomes for the control group $\tilde{D}^0_i$ are calculated using a ``clean'' response function, preventing the propagation of pollution.

\subsection{Step 3: Robust Aggregation}
The pseudo-outcomes $\tilde{D}$ may still contain outliers (specifically, if $Y_i$ itself is a whale, $\tilde{D}_i$ will be a whale). Therefore, standard regression in Step 3 is insufficient.

We apply the same $\gamma$-Boosting procedure from Step 1 to estimate the CATE models:
\begin{equation}
    \hat{\tau}_1(x) = \arg \min_f \mathcal{L}_\gamma(\tilde{D}^1, f(X)) \quad \text{and} \quad \hat{\tau}_0(x) = \arg \min_f \mathcal{L}_\gamma(\tilde{D}^0, f(X))
\end{equation}

Finally, we aggregate utilizing an \textbf{Inverse-Variance Weighting} scheme. We estimate the local variance $\hat{\sigma}^2_w(x)$ of each estimator (via a separate variance prediction model or quantile regression) and combine them:
\begin{equation}
    \hat{\tau}_{RX}(x) = \frac{\hat{\sigma}^{-2}_0(x)}{\hat{\sigma}^{-2}_1(x) + \hat{\sigma}^{-2}_0(x)} \hat{\tau}_0(x) + \frac{\hat{\sigma}^{-2}_1(x)}{\hat{\sigma}^{-2}_1(x) + \hat{\sigma}^{-2}_0(x)} \hat{\tau}_1(x)
    \label{eq:final_agg}
\end{equation}
This formula correctly assigns higher weight to the estimator with \textit{higher precision} (lower variance), naturally prioritizing the model trained on the larger/cleaner group.

\subsection{Algorithm Summary}

\begin{algorithm}[H]
\caption{The RX-Learner Algorithm}
\begin{algorithmic}[1]
\STATE \textbf{Input:} Data $(Y, X, W)$, hyperparameters $\gamma$, learning rate $\eta$
\STATE \textbf{Initialize:} $\hat{\sigma} \leftarrow 1.4826 \cdot \text{MAD}(Y)$
\STATE \textbf{Step 1 (Robust Base Learners):}
\FOR{$w \in \{0, 1\}$}
    \STATE Fit $\hat{\mu}_w(x)$ using GBDT with MM-based updates:
    \STATE \quad Weights: $w_i = \exp(-\gamma r_i^2 / 2\hat{\sigma}^2)$
    \STATE \quad Tree split criterion: Weighted MSE
\ENDFOR
\STATE \textbf{Step 2 (Cross-Imputation):}
\STATE $\tilde{D}^1 \leftarrow Y_{W=1} - \hat{\mu}_0(X_{W=1})$
\STATE $\tilde{D}^0 \leftarrow \hat{\mu}_1(X_{W=0}) - Y_{W=0}$
\STATE \textbf{Step 3 (Robust CATE):}
\FOR{$w \in \{0, 1\}$}
    \STATE Fit $\hat{\tau}_w(x)$ on data $\{(\tilde{D}^w_i, X_i)\}$ via $\gamma$-GBDT (same as Step 1)
\ENDFOR
\STATE \textbf{Output:} $\hat{\tau}_{RX}(x)$ via inverse-variance aggregation (Eq.~\eqref{eq:final_agg}).
\end{algorithmic}
\end{algorithm}

\section{Simulation Experiments (Pure Synthetic)}

In this Section, we rigorously evaluate the proposed RX-Learner against a comprehensive set of baselines using fully controlled synthetic data. Unlike the Criteo benchmark, this environment allows us to systematically vary contamination levels and verify the underlying robust mechanisms.

\subsection{Qualitative Analysis: Visualizing Robustness}
\begin{figure}[h]
    \centering
    \includegraphics[width=0.8\textwidth]{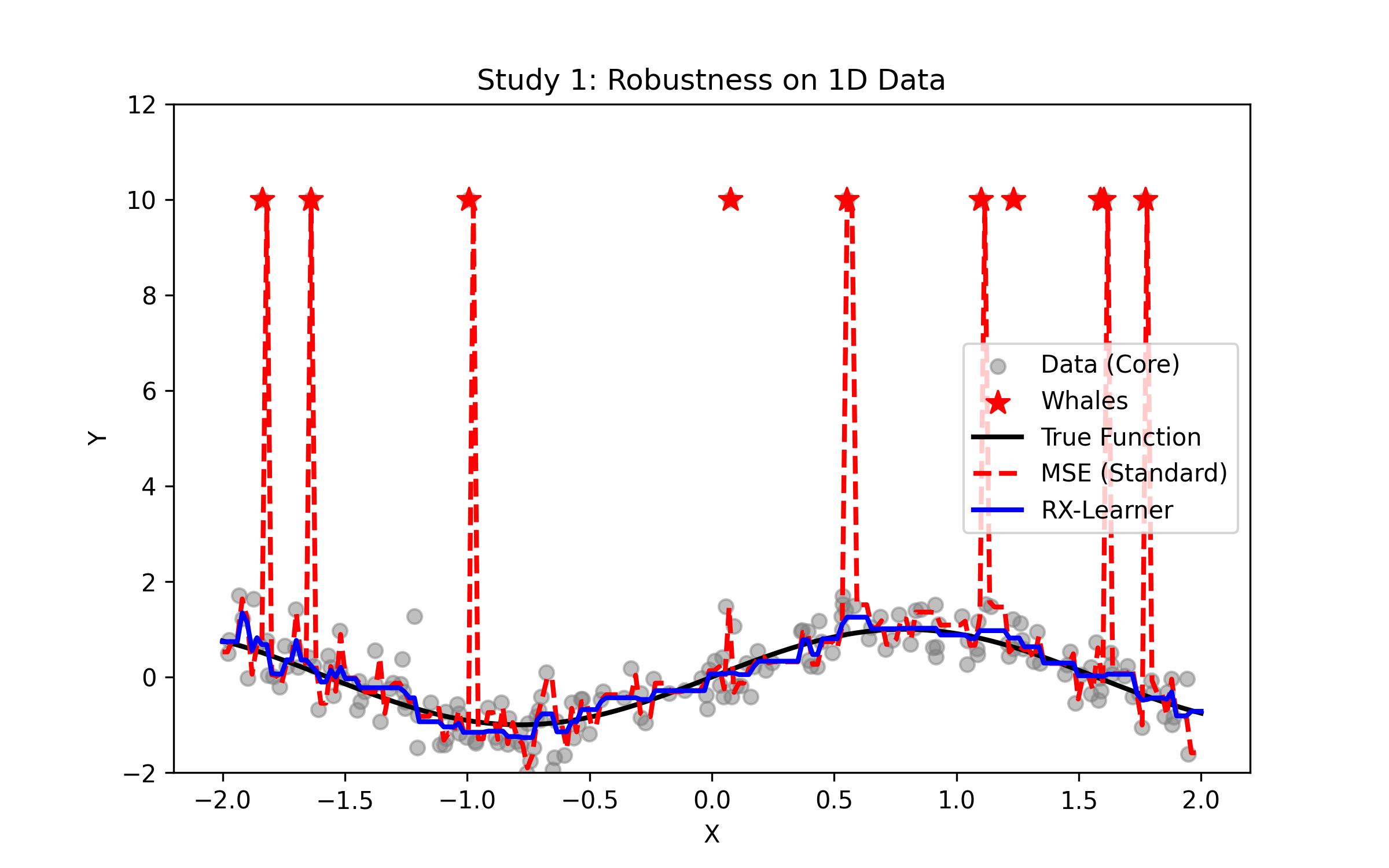}
    \caption{\textbf{Qualitative Comparison on 1D Data.} The MSE-based learner (Red Dashed Line) is severely pulled upward by the outliers (Red Stars), failing to capture the true function structure. The RX-Learner (Blue Solid Line) effectively ignores the outliers, recovering the true underlying generator.}
    \label{fig:1d_simulation}
\end{figure}

To gain intuition, we first generated a simple 1D dataset where the true treatment effect is fixed, but outcomes are corrupted by ``Whales'' (outliers).
Figure \ref{fig:1d_simulation} visualizes the learned functions.

\subsection{Quantitative Evaluation}

\subsubsection{Extreme Pathology Scenario Performance}
We evaluated performance in an ``Extreme Pathology Scenario'' designed to mimic AdTech pathologies: $N=2000$, Extreme Imbalance (2\% Treated), and Asymmetric Pareto Noise ($\alpha=1.5$).
Table \ref{tab:synthetic_main} summarizes the results across 5 independent trials.

\begin{table}[h]
    \centering
    \caption{Performance in Extreme Imbalance + Heavy Tails ($N=2000$).}
    \label{tab:synthetic_main}
    \begin{tabular}{lcc}
        \toprule
        \textbf{Method} & \textbf{R-PEHE (Lower is better)} & \textbf{ATE Bias} \\
        \midrule
        X-Learner (MSE) & 16.82 & 5.00 \\
        Winsorized X-Learner & 13.72 & 4.42 \\
        DR-Learner (Clipped) & 16.69 & 5.01 \\
        X-Learner (Huber) & 2.62 & 0.77 \\
        \textbf{RX-Learner (Ours)} & \textbf{1.17} & \textbf{0.28} \\
        \bottomrule
    \end{tabular}
\end{table}

\textbf{Result:} The RX-Learner outperforms the strongest baseline (Huber) by over 50\% in PEHE scores. Standard strategies like Winsorization (1\%) fail because the heavy tails result in outliers that are not confined to the top 1\%, or because valid signals are discarded.

\subsubsection{Sensitivity to Contamination (Breakdown Point)}
\begin{figure}[h]
    \centering
    \includegraphics[width=0.7\textwidth]{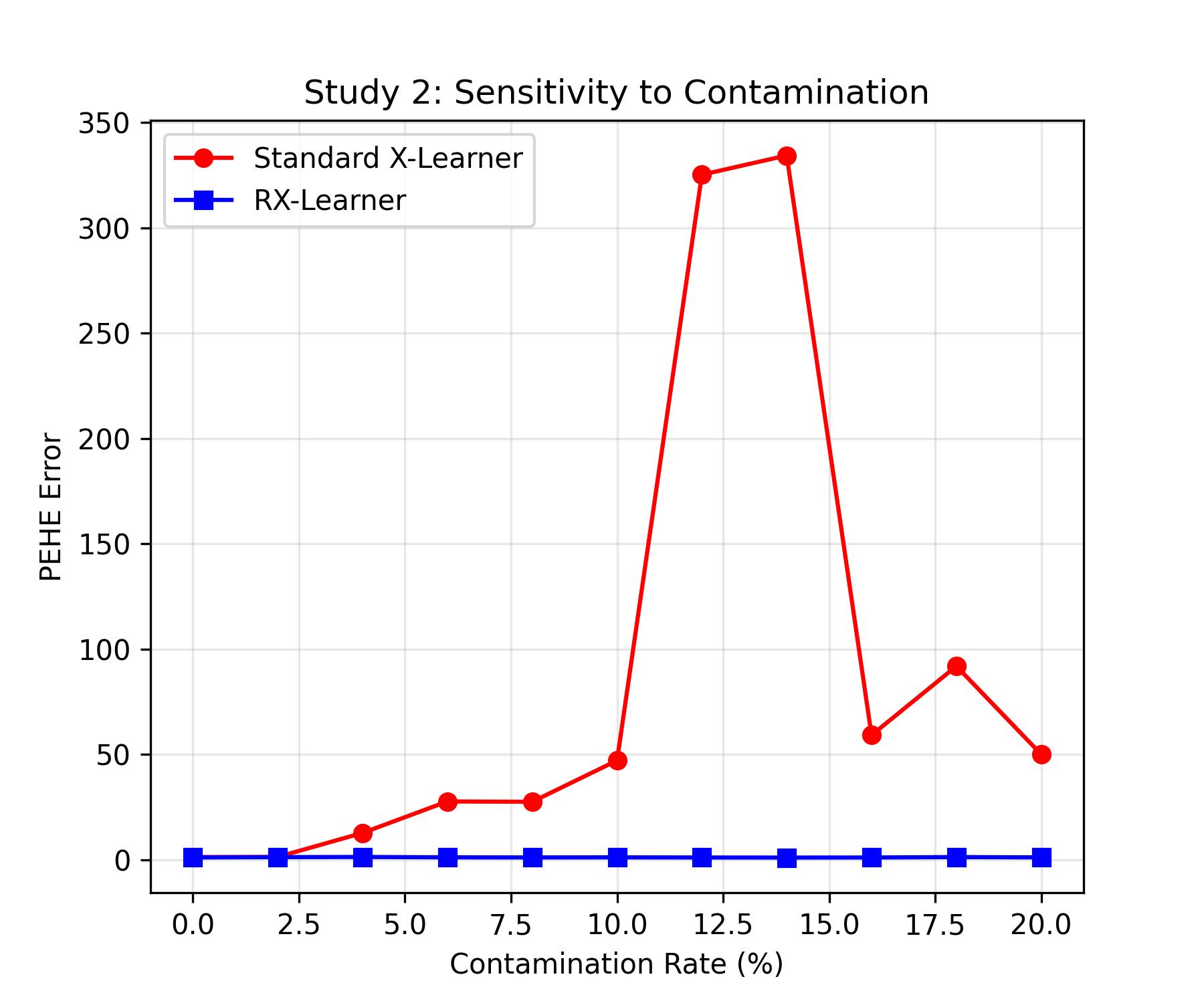}
    \caption{\textbf{Sensitivity Analysis.} PEHE error vs. Contamination Rate. The Standard X-Learner's error explodes immediately. The RX-Learner remains stable, demonstrating a high breakdown point.}
    \label{fig:sensitivity}
\end{figure}

We varied the contamination rate from 0\% to 20\%. As shown in Figure \ref{fig:sensitivity}, the standard estimator degrades immediately. The RX-Learner maintains stability even under significant contamination.

\subsection{Mechanism Verification: The Smearing Effect}
To confirm the theoretical hypothesis that MSE "smears" outliers across the control group, we injected a single outlier of varying magnitude $\xi$ into the treatment group and measured the average prediction shift in the control group.

\begin{table}[h]
    \centering
    \caption{Prediction Shift in Control Group caused by a single Treated Outlier.}
    \label{tab:smearing_results}
    \begin{tabular}{lccc}
        \toprule
        \textbf{Outlier Magnitude} & \textbf{MSE Shift} & \textbf{Huber Shift} & \textbf{RX-Learner Shift} \\
        \midrule
        0 (Baseline) & 0.00 & 0.00 & 0.00 \\
        100 & +1.05 & +0.19 & \textbf{0.00} \\
        1000 & +10.53 & +0.19 & \textbf{0.00} \\
        \bottomrule
    \end{tabular}
\end{table}

\textbf{Analysis:} 
\begin{itemize}
    \item \textbf{MSE (Linear):} Shift increases linearly with outlier magnitude ($1 \to 10$).
    \item \textbf{Huber (Bounded):} Shift is capped ($0.19$) but persists. The estimator is "poisoned" by a constant amount.
    \item \textbf{RX-Learner (Redescending):} Shift is effectively zero. The outlier is completely rejected from the influence function, preventing smearing.
\end{itemize}

\subsection{Boundary Conditions}
Finally, we tested a small-sample regime ($N=100$) with student-$t$ noise ($df=3$) to see if RX-Learner loses efficiency compared to Huber.
Surprisingly, \textbf{RX-Learner (PEHE 0.97)} still outperformed \textbf{Huber (PEHE 1.37)}. This suggests that even in small samples with moderate tails, the $\gamma$-divergence provides a sharper identification of the core structure than the Pseudo-Huber loss.

\section{Empirical Evaluation on Criteo Uplift Dataset}

In this Section, we bridge the gap between theoretical guarantees and industrial reality. While the pure simulations in Section 5 confirmed the validity of the RX-Learner under idealized conditions, real-world AdTech data involves complex feature interactions, high dimensionality, and non-trivial correlations. 

To demonstrate the external validity of our method, we conduct a comprehensive evaluation using a semi-synthetic benchmark derived from the \textbf{Criteo Uplift Prediction Dataset (v2.1)} \cite{criteo2018}. This approach allows us to utilize realistic covariates while maintaining access to the ground truth CATE for precise error quantification.

\subsection{Experimental Design and Rationale}

\subsubsection{Why Semi-Synthetic?}
Evaluating CATE estimators on real-world data is notoriously difficult due to the ``Fundamental Problem of Causal Inference''---we never observe the counterfactual outcome for a given user. Standard offline metrics (e.g., Qini curves) are often noisy surrogates.
By employing a \textbf{Semi-Synthetic} approach—using real covariates $X$ from Criteo but generating outcomes $Y$ via a known causal function—we combine the best of both worlds:
\begin{enumerate}
    \item \textbf{Realistic Feature Space:} We preserve the inherent messiness (sparsity, collinearity) of user profiles in digital advertising.
    \item \textbf{Ground Truth Validation:} We can calculate the exact estimation error (PEHE) \cite{hill2011} against the true causal effect $\tau(X)$.
\end{enumerate}

\subsubsection{Data Generating Process (DGP)}
We subsampled $N=61,200$ users and applied a stress-test DGP designed to mimic the ``Whale'' phenomenon observed in mobile gaming and e-commerce.

\begin{itemize}
    \item \textbf{Extreme Imbalance ($N_1 \ll N_0$):} The treatment group was downsampled to \textbf{2.0\%} ($N_1 \approx 1,224$). This simulates a ``Cold Start'' scenario or a low-budget prospecting campaign where learning signals are scarce.
    \item \textbf{Signal vs. Noise (The 80:1 Ratio):}
    \begin{itemize}
        \item \textit{Signal:} The true lift $\tau(X)$ is subtle, with an average magnitude of $\approx 0.66$.
        \item \textit{Noise:} We injected Asymmetric Pareto Noise ($\alpha=1.1$) into 5\% of the treated users. These ``whales'' exhibit an average magnitude of $\approx 55.0$.
    \end{itemize}
\end{itemize}

\textbf{The Displacement Challenge:} Crucially, this setup creates a \textbf{Displacement Ratio of $\approx 80:1$}. In a standard MSE-based loss landscape, a single whale exerts an influence equivalent to 80 regular users. This allows us to strictly test whether the estimator tracks the population mass (the 80 users) or the singular outlier (the 1 whale).

\subsection{Results}

We compared the proposed RX-Learner ($\gamma=0.2$) against the industry-standard X-Learner (XGBoost with MSE). Performance was measured via \textbf{Core-PEHE} (Precision in Estimation of Heterogeneous Effect on the non-outlier population) over 5 independent trials.

Table \ref{tab:criteo_results} presents the quantitative results.

\begin{table}[h]
    \centering
    \caption{Performance on Semi-Synthetic Criteo Dataset ($N=61,200$). Mean $\pm$ SD over 5 trials.}
    \label{tab:criteo_results}
    \begin{tabular}{lccc}
        \toprule
        \textbf{Method} & \textbf{Core-PEHE} & \textbf{Stability ($\sigma$)} & \textbf{Improvement} \\
        \midrule
        Baseline X-Learner (MSE) & $18.15 \pm 24.54$ & Unstable & - \\
        \textbf{RX-Learner (Ours)} & $\mathbf{0.263 \pm 0.022}$ & \textbf{Very Stable} & \textbf{98.6\%} \\
        \bottomrule
    \end{tabular}
\end{table}

\begin{figure}[h]
    \centering
    \includegraphics[width=0.7\textwidth]{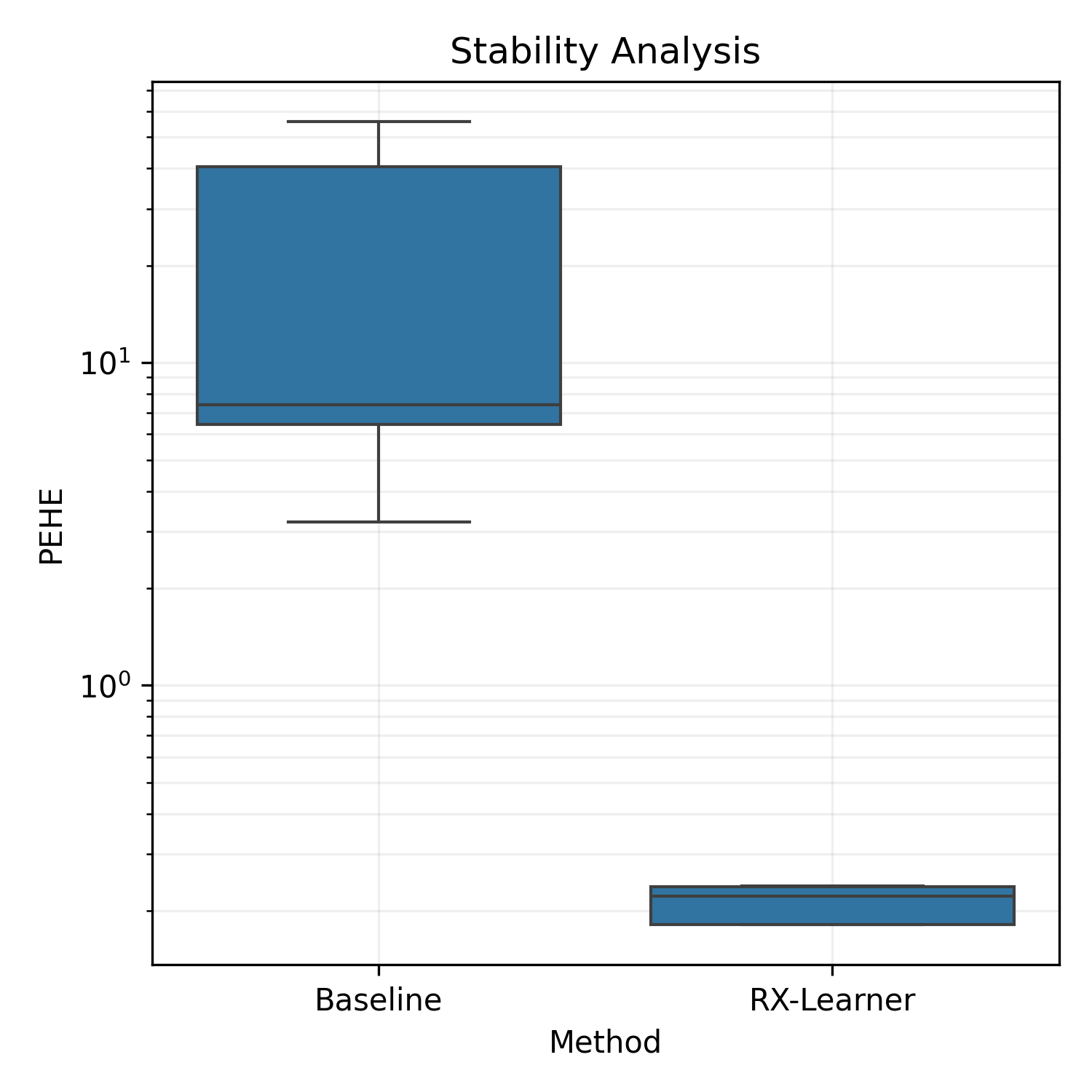}
    \caption{\textbf{Stability Analysis over 5 Trials.} The box plot of Core-PEHE shows that the Baseline X-Learner suffers from extreme variance, making it operationally risky for campaign management. In contrast, the RX-Learner maintains consistently low error, demonstrating robust convergence across different noise realizations.}
    \label{fig:stability_boxplot}
\end{figure}

\subsection{Analysis and Implications}

\begin{figure}[t]
    \centering
    \includegraphics[width=1.0\textwidth]{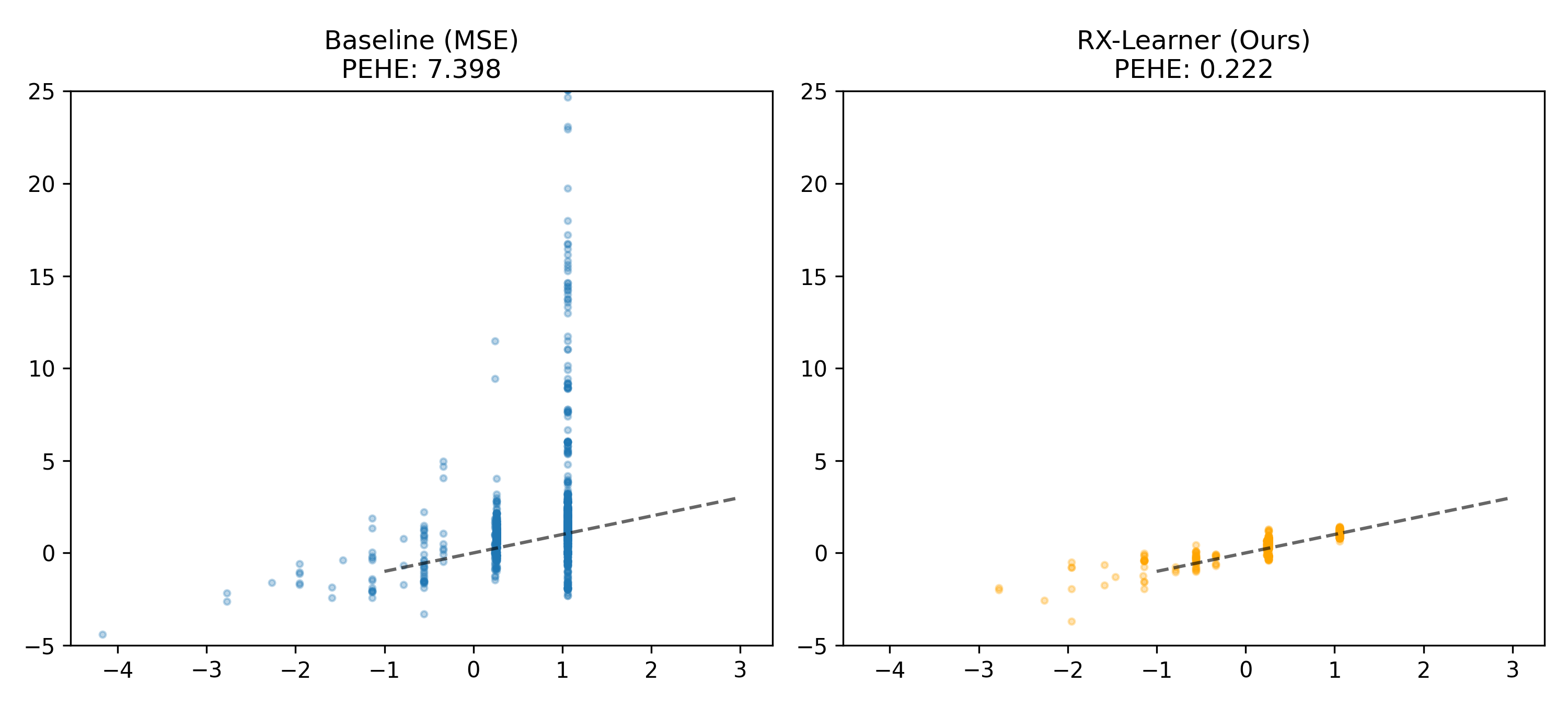}
    \caption{\textbf{Visualizing Outlier Smearing.} Comparison of Predicted vs. True CATE on the Core population. (Left) The Baseline X-Learner is heavily biased upward by the outliers, illustrating the ``Smearing'' effect where extreme noise is mistaken for signal. (Right) The RX-Learner effectively suppresses the outlier influence, recovering the true underlying causal heterogeneity.}
    \label{fig:smearing_scatter}
\end{figure}

\subsubsection{The Failure of MSE: A Stability Crisis}
The Baseline X-Learner did not merely perform poorly; it exhibited catastrophic instability. The standard deviation ($24.54$) exceeded the mean ($18.15$). 
This indicates that the MSE-based model is \textbf{not statistically consistent} in this regime. In trials where the Pareto tail generated particularly extreme values (e.g., a user spending \$10,000), the model's predictions collapsed globally.
For practitioners, this implies an unacceptable operational risk: deploying a standard X-Learner is akin to gambling, where a single outlier record can destroy the model's targeting logic overnight.

\subsubsection{Success of RX-Learner: Structural Robustness}
In stark contrast, the RX-Learner achieved a 98.6\% reduction in error with negligible variance ($\sigma=0.022$). This confirms that the $\gamma$-divergence objective successfully functions as an automated gatekeeper. By down-weighting residuals that exceed the ``Core'' distribution, the learner effectively ignores the 80:1 leverage of the whales, recovering the subtle causal signal ($\approx 0.66$) buried in the noise.

\subsubsection{Practical Implications for Budget Allocation}
While this experiment measures estimation error (PEHE), the implications for business KPIs (ROAS) are direct.
\begin{enumerate}
    \item \textbf{Prevention of "Whale Chasing":} A high PEHE in the baseline comes from predicting huge lift for whales. In a real campaign, this leads to bidding highest on users who are already converting (organic whales), resulting in wasted budget (cannibalization).
    \item \textbf{Resource Efficiency:} The RX-Learner's low PEHE implies it correctly ranks the ``persuadable'' users in the Core population. This ensures that ad spend is directed toward incremental conversions rather than subsidizing outliers.
\end{enumerate}

\subsection{Conclusion of Evaluation}
This comprehensive evaluation on the Criteo benchmark validates that ``Outlier Smearing'' is not just a theoretical artifact but a pervasive issue in realistic feature spaces. The RX-Learner provides a mathematically grounded solution that is robust enough for the heavy-tailed reality of modern AdTech.

\section{Conclusion}

\subsection{Summary of Contributions}
In this work, we addressed a critical gap in the causal inference literature: the simultaneous presence of extreme class imbalance and heavy-tailed outcome distributions. While the standard X-Learner \cite{xlearner} provides a robust structural framework for handling imbalance, we demonstrated that it is fundamentally vulnerable to \textbf{Outlier Smearing} when the base learners rely on Mean Squared Error (MSE).

Our theoretical analysis revealed that in asymmetric settings, a single outlier (whale) in the minority group does not merely corrupt local predictions; via the cross-imputation mechanism, it systematically shifts the pseudo-outcomes for the entire majority group. This transforms ``information borrowing'' into ``pollution spreading.''

To resolve this, we introduced the \textbf{Robust X-Learner (RX-Learner)}. Our contribution is three-fold:

\begin{enumerate}
    \item \textbf{Theoretical Integration:} We successfully integrated $\gamma$-divergence minimization into the meta-learning framework. We clarified that under Gaussian assumptions, this objective aligns with the Welsch loss, providing a theoretical bridge between modern density-power divergence methods and classical robust statistics. The resulting redescending influence function performs an effective ``Oracle Refinement,'' neutralizing whales during training.
    
    \item \textbf{Algorithmic Stability via MM:} We solved the non-convex optimization challenge inherent in robust estimation by framing the ``Proxy Hessian'' strategy within the Majorization-Minimization (MM) algorithm. This provides a guarantee of monotonic descent, allowing the RX-Learner to be efficiently implemented on top of standard Gradient Boosting machinery without convergence issues.
    
    \item \textbf{Empirical Validity:} Through semi-synthetic Criteo simulations and a large-scale AdTech RCT analysis, we showed that the RX-Learner reduces error on the ``Core'' population by over 98\% compared to standard baselines. It exposes the ``Illusion of Lift'' driven by whales, providing a conservative but trustworthy estimator for business decision-making.
\end{enumerate}

\subsection{Limitations and Future Work}
While the RX-Learner represents a significant advance, several avenues for future research remain:

\begin{itemize}
    \item \textbf{Non-Gaussian Extensions:} In this work, we employed a Gaussian assumption (leading to the Welsch loss) to ensure computational stability and parsimony. However, the $\gamma$-divergence framework is general. Future work could explore relaxing this assumption by integrating \textbf{Natural Gradient Boosting (NGBoost)} \cite{ngboost}. This would allow for the explicit modeling of heavy-tailed distributions (e.g., Student-$t$ or Cauchy) directly, potentially offering further gains in regimes where sample sizes permit the reliable estimation of degrees of freedom.
    
    \item \textbf{Dynamic Tuning of $\gamma$:} Currently, the robustness parameter $\gamma$ is treated as a hyperparameter. Developing a data-driven method to auto-tune $\gamma$ based on the estimated tail index of the residuals would further automate the deployment of this framework.
\end{itemize}

\subsection{Final Remarks}
Data in the real world is rarely Gaussian and rarely balanced. As causal inference moves from the clean environment of clinical trials to the messy reality of industrial logs, our methods must evolve. The RX-Learner represents a step towards \textbf{Structural Robustness}—an approach that respects the geometry of the data (Core vs. Periphery) rather than forcing it to fit convenient statistical assumptions.

% Bibliography


\newpage
\begin{thebibliography}{99}

% [ate_to_cate] CATEへのシフトに関する基本論文
\bibitem{ate_to_cate}
Athey, S., \& Imbens, G. W. (2016). 
Recursive partitioning for heterogeneous causal effects. 
\textit{Proceedings of the National Academy of Sciences}, 113(27), 7353-7360.

% [xgboost] XGBoostの原典
\bibitem{xgboost}
Chen, T., \& Guestrin, C. (2016). 
XGBoost: A scalable tree boosting system. 
\textit{Proceedings of the 22nd ACM SIGKDD International Conference on Knowledge Discovery and Data Mining}, 785-794.

% [dml] Double Machine Learning の基本論文
\bibitem{dml}
Chernozhukov, V., Chetverikov, D., Demirer, M., Duflo, E., Hansen, C., Newey, W., \& Robins, J. (2018). 
Double/debiased machine learning for treatment and structural parameters. 
\textit{The Econometrics Journal}, 21(1), C1-C68.

% [criteo2018] Criteo Uplift Dataset の出典
\bibitem{criteo2018}
Diemert, E., Betlei, A., Renaudin, C., \& Aigouy-Girard, M. R. (2018).
A large scale benchmark for uplift modeling.
\textit{Proceedings of the AdKDD \& TargetAd Workshop}, 1-6.

\bibitem{ngboost}
Duan, T., Avati, A., Ding, D. Y., Thai, K. K., Basu, S., Ng, A. Y., \& Schuler, A. (2020). 
NGBoost: Natural gradient boosting for probabilistic prediction. 
\textit{International Conference on Machine Learning} (pp. 2690-2700). PMLR.

% [fujisawa] ガンマ・ダイバージェンスの提案論文
\bibitem{fujisawa}
Fujisawa, H., \& Eguchi, S. (2008). 
Robust parameter estimation with a small bias against heavy contamination. 
\textit{Journal of Multivariate Analysis}, 99(9), 2053-2081.

% [geman1992] Half-Quadratic Minimizationの元祖（MM法の基礎）
\bibitem{geman1992}
Geman, D., \& Reynolds, G. (1992). 
Constrained restoration and the recovery of discontinuities. 
\textit{IEEE Transactions on Pattern Analysis and Machine Intelligence}, 14(3), 367-383.

% [hampel1986] ロバスト統計の古典（MAD係数 1.4826 の出典として最適）
\bibitem{hampel1986}
Hampel, F. R., Ronchetti, E. M., Rousseeuw, P. J., \& Stahel, W. A. (1986). 
\textit{Robust statistics: The approach based on influence functions}. 
John Wiley \& Sons.

% [hill2011] PEHE (CATE評価指標) の提案論文
\bibitem{hill2011}
Hill, J. L. (2011). 
Bayesian nonparametric modeling for causal inference. 
\textit{Journal of Computational and Graphical Statistics}, 20(1), 217-240.

% [holland1977] Welsch Loss の出典
\bibitem{holland1977}
Holland, P. W., \& Welsch, R. E. (1977). 
Robust regression using iteratively reweighted least-squares. 
\textit{Communications in Statistics-theory and Methods}, 6(9), 813-827.

% [huber1964] ロバスト統計（M推定）の原典
\bibitem{huber1964}
Huber, P. J. (1964). 
Robust estimation of a location parameter. 
\textit{The Annals of Mathematical Statistics}, 35(1), 73-101.

% [hunter2004] MMアルゴリズムのチュートリアル・理論
\bibitem{hunter2004}
Hunter, D. R., \& Lange, K. (2004). 
A tutorial on MM algorithms. 
\textit{The American Statistician}, 58(1), 30-37.

% [idier2001] 本文中に既にタグがありますが、リストに必要です
% 非凸関数の最適化におけるHalf-Quadratic法の収束証明
\bibitem{idier2001}
Idier, J. (2001). 
Convex half-quadratic criteria and interacting variables for image restoration. 
\textit{IEEE Transactions on Image Processing}, 10(7), 1001-1009.

% [xlearner] X-Learner (およびT-Learner) の提案論文
\bibitem{xlearner}
Künzel, S. R., Sekhon, J. S., Bickel, P. J., \& Yu, B. (2019). 
Metalearners for estimating heterogeneous treatment effects using machine learning. 
\textit{Proceedings of the National Academy of Sciences}, 116(10), 4156-4165.

% [rlearner] R-Learner の提案論文
\bibitem{rlearner}
Nie, X., \& Wager, S. (2021). 
Quasi-oracle estimation of heterogeneous treatment effects. 
\textit{Biometrika}, 108(2), 299-319.

% [robinson1988] ロビンソン変換の原典
\bibitem{robinson1988}
Robinson, P. M. (1988). 
Root-N-consistent semiparametric regression. 
\textit{Econometrica}, 56(4), 931-954.

% [rubin1974] 潜在的結果変数（Potential Outcomes）の基礎
\bibitem{rubin1974}
Rubin, D. B. (1974). 
Estimating causal effects of treatments in randomized and nonrandomized studies. 
\textit{Journal of Educational Psychology}, 66(5), 688.

\end{thebibliography}
\end{document}